\documentclass[0.5pt,letterpaper]{article}
\usepackage{fullpage}

\usepackage[utf8]{inputenc} 
\usepackage[T1]{fontenc}    
\usepackage{hyperref}       
\usepackage{url}            
\usepackage{hyperref}
\usepackage{booktabs}       
\usepackage{amsfonts}       
\usepackage{nicefrac}       
\usepackage{microtype}      

\usepackage{amsmath}
\usepackage{amssymb}
\usepackage{wasysym}
\usepackage{multirow}
\usepackage[dvipsnames]{xcolor}
\usepackage{amsthm}
\usepackage{bbm}
\usepackage{enumerate}
\usepackage{footmisc}

\usepackage{amsmath}
\usepackage{amsfonts}
\usepackage{graphicx}

\usepackage{colortbl}
\usepackage{cancel}
\usepackage{multirow}
\usepackage{relsize}
\usepackage{algorithm}
\usepackage[noend]{algpseudocode}
\usepackage{forloop}
\usepackage{bm}
\usepackage{mathtools}
\usepackage{subfigure}
\usepackage{booktabs}
\usepackage{times}
\usepackage{etoolbox}
\usepackage{enumitem}
\usepackage{xfrac}
\usepackage[group-separator={,}]{siunitx}
\usepackage[skins]{tcolorbox}

\usepackage{accents}

\newcommand\sbullet[1][.5]{\mathbin{\vcenter{\hbox{\scalebox{#1}{$\bullet$}}}}}
\renewcommand{\dot}[1]{    {\stackrel{\scriptscriptstyle \sbullet[0.42]}{#1}}    }

\DeclareMathOperator{\argmin}{argmin}

\newcommand{\new}{\text{\tiny new}}
\DeclareMathOperator{\sign}{sgn}

\newcommand{\zero}{\bm{0}}

\renewcommand{\u}{\bm{u}}
\renewcommand{\v}{\bm{v}}
\newcommand{\bnu}{\bm{\nu}}
\newcommand{\g}{\bm{g}}
\newcommand{\gt}{\tilde{\g}}
\newcommand{\ema}{\bm{m}}
\newcommand{\s}{\bm{s}}

\newcommand{\p}{\bm{p}}

\newcommand{\w}{\bm{w}}
\newcommand{\x}{\bm{x}}
\newcommand{\thet}{\bm{\theta}}
\newcommand{\thett}{\tilde{\thet}}

\newcommand{\betat}{\tilde{\beta}}

\newcommand{\X}{\mathcal{X}}
\newcommand{\RR}{\mathbb{R}}
\newcommand{\one}{\bm{1}}

\newcommand{\partilde}{\widetilde{\nabla}}

\makeatletter
\DeclareRobustCommand{\emat}{\mathbin{\mathpalette\emat@@\relax}}
\newcommand{\emat@@}[2]{%
  \vbox{\offinterlineskip
    \sbox\z@{$\m@th#1\ema$}%
    \ialign{%
      \hfil##\hfil\cr
      $\m@th#1{}_{\sim}\kern-\scriptspace$\cr
      \noalign{\kern+0.01\ht\z@}
      \box\z@\cr
    }%
  }%
}
\makeatother

\makeatletter
\DeclareRobustCommand{\wt}{\mathbin{\mathpalette\wt@@\relax}}
\newcommand{\wt@@}[2]{%
  \vbox{\offinterlineskip
    \sbox\z@{$\m@th#1\w$}%
    \ialign{%
      \hfil##\hfil\cr
      $\m@th#1{}_{\sim}\kern-\scriptspace$\cr
      \noalign{\kern+0.01\ht\z@}
      \box\z@\cr
    }%
  }%
}
\makeatother

\definecolor{darkgreen}{rgb}{0.0, 0.5, 0.0}
\definecolor{darkpink}{rgb}{0.72254902,  0.05588235,  0.44019608}

\usepackage{bm}
\usepackage{amssymb}
\usepackage{xcolor}
\usepackage{pgfplots}

\definecolor{NiceOrange}{RGB}{0, 0, 0}
\definecolor{NiceBlue}{RGB}{0, 0, 0}
\definecolor{NiceGreen}{RGB}{10, 10, 10}
\definecolor{NiceOrange2}{RGB}{0, 0, 0}

\date{}

\title{Step-size Adaptation Using Exponentiated Gradient Updates}

\author{%
  Ehsan Amid\footnote{A shorter version of this paper appeard in Workshop on ``Beyond first-order methods in ML systems'' at the 37th International Conference on Machine Learning (ICML), 2020: \href{https://users.soe.ucsc.edu/~eamid/paps/funnel_ws.pdf}{\texttt{https://users.soe.ucsc.edu/\textasciitilde eamid/paps/funnel\_ws.pdf}}}\qquad Rohan Anil\qquad Christopher Fifty\qquad Manfred K. Warmuth \\
  Google Research, Brain Team \\
  Mountain View, CA \\
  \texttt{\{eamid, rohananil, cfifty, manfred\}@google.com} \\
}

\begin{document}

\maketitle

\begin{abstract}
    Optimizers like Adam and AdaGrad have been very successful in training large-scale neural networks. Yet, the performance of these methods is heavily dependent on a carefully tuned learning rate schedule. We show that in many large-scale applications, augmenting a given optimizer with an adaptive tuning method of the step-size greatly improves the performance. More precisely, we maintain a global step-size scale for the update as well as a gain factor for each coordinate. We adjust the global scale based on the alignment of the average gradient and the current gradient vectors. A similar approach is used for updating the local gain factors. This type of step-size scale tuning has been done before with gradient descent updates. In this paper, we update the step-size scale and the gain variables with exponentiated gradient updates instead. Experimentally, we show that our approach can achieve compelling accuracy on standard models without using any specially tuned learning rate schedule. We also show the effectiveness of our approach for quickly adapting to distribution shifts in the data during training.
\end{abstract}
\section{Introduction}

Successful training of large neural networks heavily depends on the choice of a good optimizer as well as careful tuning of the hyperparameters such as the learning rate, momentum, weight decay, etc. Among the hyperparameters, the learning rate schedule is one of the most important elements for achieving the best accuracy. In many cases, the schedule consists of an initial ramp-up followed by a number of stair-case decays throughout the training phase. The schedule is typically tailored manually for the problem and requires re-training the model numerous times.

The most recently proposed optimizers for deep neural networks (RMSProp~\cite{rmsprop}, AdaGrad~\cite{adagrad}, Adam~\cite{adam}, etc.) have been based on adapting the gradient via a (diagonal) pre-conditioner matrix. However, these techniques still require a carefully tuned learning rate schedule to achieve the optimal performance. Furthermore, little has been done for adapting the per-coordinate or the overall step-size. The common idea among the few available approaches ~\cite{dbd,rprop,nicol,hypergrad} is to move faster along directions that are making progress and punish those that alternate often. However, for the previous methods, 1) the formulation is not general enough to be compatible with different optimizers;  and 2), the update equations are based on inferior heuristics or in some cases use the incorrect gradients~\cite{nicol} (as will be discussed later).

In this paper, we aim to unify such approaches in a more rigorous manner. Concretely, we propose an abstraction in the form of a momentum update by passing the pre-conditioned gradient, proposed by an arbitrary internal optimizer to the meta algorithm. We then make the step-size adaptive by introducing the following hyperparameters: one overall step-size scale as well as local gain factors for each coordinate. The scale and gains are non-negative and trained using the Unnormalized Exponentiated Gradient (EGU) updates~\cite{eg}. As a brief introduction, EGU minimizes a function $f$ by adding a relative entropy (a.k.a. Kullback-Leibler) divergence~\cite{kl} as an \emph{inertia} term. The goal of adding the inertia term is to keep the updated parameters $\thet^{t+1}$ close to the previous parameter $\thet^t$ at step $t$:
\begin{equation}
\label{eq:min}
\thet^{t+1} = \argmin_{\,\thett \succeq \zero}\big\{\sfrac{1}{\eta}\, D_{\text{\tiny RE}}(\thett, \thet^t) + f(\thett)\big\}\,,
\end{equation}
where $\eta > 0$ is a learning rate parameter and
\[
D_{\text{\tiny RE}}(\u, \v) = \sum_i\big(u_i \log\frac{u_i}{v_i} - u_i + v_i\big)\, .
\]
The update multiplies each parameter by an exponentiated gradient factor:
\begin{equation}
\label{eq:egu}
\thet^{t+1} = \thet^t \odot \exp\big(-\eta\, \nabla_{\thet} f(\thet^t)\big)\, , \qquad\text{(EGU)}
\end{equation}
where $\nabla_{\thet} f(\thet^t)$ denotes the gradient of the objective function $f(\thet)$ w.r.t. $\thet$ evaluated at $\thet^t$ and $\odot$ denotes element-wise product\footnote{The exact minimization of \eqref{eq:min} uses the gradient $\nabla_{\thet} f(\thet^{t+1})$ which is then approximated by $\nabla_{\thet} f(\thet^t)$ in the EGU update. More on this later.}. The multiplicative form of the update ensures $\thet^{t+1} \succcurlyeq \zero$ at any time.

The properties of the EGU update have been studied extensively in the online learning literature~\cite{eg,hedge,percwin,singer,pnorm,warmuth2008}. Specifically, it has been shown that the EGU update converges significantly faster than gradient descent in cases when only a small subset of the dimensions are relevant~\cite{winnow,matrixwinnow,wincolt}. As a result, EGU is extremely efficient in discovering the relevant dimensions while immediately damping the irrelevant ones. Also, the EGU update naturally maintains the non-negativity of the parameters. Finally, the multiplicative form of the update allows exploring a wider range of values for the parameters more rapidly.

\paragraph{Contributions}
In this paper, we build upon the ideas of~\cite{dbd,nicol,hypergrad} and introduce a unified approach for step-size adaptation based on the unnormalized exponentiated gradient updates. The main goal of the paper is to revisit the previously developed ideas in different domains and show cases where such updates are effective on large deep neural networks. In summary:
\vspace{-0.1cm}
\begin{itemize}[leftmargin=4mm]
    \item We introduce a step-size adaptation framework which introduces per-coordinate gains as well a step-size scale for the update. We apply the EGU updates on these hyperparameters.
    \item Our formulation is versatile and accepts any pre-conditioned gradient by an adaptive gradient optimizer as input. Thus, it can be coupled with a wide range of commonly used optimizers.
    \item We show extremely promising use cases for our adaptive step-size method and discuss potential extensions of such adaptive methods.
    \item We show the efficacy of our method by conducting an extensive set of experiments on large-scale neural network on benchmark datasets and publish the code for reproducibility at:~\url{https://users.soe.ucsc.edu/~eamid/funnel.html}.
\end{itemize}

\subsection{Related Work}
Introducing per-coordinate gains dates back to the Delta-Bar-Delta (DBD) method~\cite{dbd} where the set of gains are adaptively updated using the sign agreements of the current gradient and an exponential running average (EMA) of the past gradients. DBD uses a mixture of additive and multiplicative updates where the gains for the coordinates with agreeing gradient signs are increased by a constant amount and the remaining gains are damped by a multiplicative factor. Later, these ideas were extended to different settings~\cite{survey-dbd,sutton}. More relevantly, a local gain adaptation method was introduced in~\cite{nicol} where the goal was to update the gains using the EGU updates. However, the wrong gradient term (gradient w.r.t. $\log$-gains instead of gradient w.r.t. gains) was used on the final updates, i.e.
\[
\thet^{t+1} = \thet^t \odot \exp\big(-\eta\, \nabla_{\log\thet} f(\thet^t)\big) = \thet^t \odot \exp\big(-\eta\, \thet \odot \nabla_{\thet} f(\thet^t)\big)\, .\qquad\text{(incorrect EGU)}
\]
Thus this update simply amounts to GD updates on the $\log$-gains of the parameters followed by an exponentiation. A more recent approach~\cite{precond} uses a PSD pre-conditioner gain matrix, which is trained with gradient descent updates on the factorized from. In this paper, we only consider the diagonal pre-conditioner and leave the extensions to the matrix case to the future work.

In terms of step-size scale adaptation, the more recent Hypergradient Descent method~\cite{hypergrad} introduces a single adaptive scale parameter into the existing optimizers and applies gradient descent updates on the scale. The authors also propose a multiplicative form of the update which is proportional to the value of the scale parameter. We show that this multiplicative form of the update is a crude approximation of the EGU update, used in our method.


\section{A Meta Algorithm for Adaptive Step-size}
Our adaptive learning rate meta algorithm accepts as input a pre-conditioned gradient from an \emph{internal optimizer} $\mathcal{D}$ based on the value of the current weight parameters $\w^t$ (and possibly, the cumulative statistic of all the previous steps). The internal optimizer only interacts with the meta algorithm via the values of the parameter and its role is to only generates the pre-conditioned gradients. Let $\gt^t \coloneqq \partilde_{\w} L(\w^t|\, \X^t)$ denote the pre-conditioned gradient generated by the internal optimizer at step $t$ using the batch of data $\X^t$. Our \textbf{``Funnel''} meta algorithm applies the following update:
\begin{equation}
\label{eq:egdd-main}
\boxed{
    \begin{split}
        \bnu^{t+1} & = \mu\, \bnu^t + \eta\, \big(\p^{t+1} \odot \tilde{\g}^t\big)\\
        \w^{t+1} & = \w^t - s^{t+1}\, \bnu^{t+1}\, .
    \end{split}
    }
\end{equation}
The update~\eqref{eq:egdd-main} in fact resembles a heavy-ball momentum update\footnote{A similar abstraction can be applied using Nesterov momentum~\cite{nesterov}.} with \emph{base learning rate} $\eta$ and \emph{momentum} hyperparameter $\mu$, with the addition of two extra elements: 1) a non-negative {per-coordinate gain} vector $\p \succcurlyeq \zero$ which component-wise multiplies the pre-conditioned gradient vector, and 2) a non-negative step-size \emph{scale} $s \geq 0$ which scales the final step. The goal of the gain hyperparameter $\p$ is to independently scale each coordinate of the pre-conditioned gradient vector. On the other hand, the step-size scale $s$ adjust the final update that is added to the parameters.

The gain as well as the scale hyperparameters are updated along with the weights at each step. Since both gains and scale are non-negative, a natural choice for the update is the EGU update~\eqref{eq:egu}. The multiplicative form of EGU updates allows these hyperparameters to effectively adapt to the dynamics of training by changing in a wider range of values more rapidly. The updates are applied by first calculating the gradient of the loss w.r.t. each hyperparameter. That is,
\begin{equation*}
\begin{split}
\label{eq:dbd-gain-up}
\nabla_{\p} L(\w^{t}|\, \X) & = -\nabla_{\w} L(\w^{t}|\, \X) \odot s^{t+1}\,\frac{\partial}{\partial \p}\big(\underbrace{\mu\, \bnu^t + \p \odot \nabla_{\w} L(\w^{t-1}|\, \X)}_{\bnu^{t}}\big)\big)\\[-2mm]
    & \approx  -s^{t+1}\nabla_{\w} L(\w^{t}|\, \X) \odot \nabla_{\w} L(\w^{t-1}|\, \X)\, .
    \end{split}
\end{equation*}
where we omit the long-term dependencies on the gain hyperparameters. Let $\g^t \coloneqq \nabla_{\w} L(\w^t|\, \X^t)$ denote the gradient of the loss using the batch of data $\X^t$. We also remove the $s^{t+1}$ term from the gradient to reduce the inter-dependency of the gains and the scale. Thus, the exponentiated gradient gain update becomes
\begin{equation}
\label{eq:gain-pre}
    \bm{p}^{t+1} = \bm{p}^t \odot \exp\big(\gamma_p\,\g^t \odot \tilde{\g}^{t-1} \big)\, ,
\end{equation}
where $\gamma_p \geq 0$ is the gain learning rate hyperparameter. Notice that update~\eqref{eq:gain-pre} depends on the value of the gradient on the current batch $\X^t$ and the value of the pre-conditioned gradient at the previous batch $\X^{t-1}$. To account for the stochasticity of the gradients due to different batches of data, we replace the second term in the gradient by an exponential moving average (EMA) of all the past pre-conditioned gradients, that is,
\begin{equation}
\label{eq:gain-up}
    \p^{t+1} = \p^t \odot \exp\big(\gamma_p\,\g^t \odot \emat^t \big)\, ,
\end{equation}
where
\[
\ema^{t+1} = \beta\, \ema^t + (1 - \beta)\, \gt^t \text{ \,\, and \,\, } \emat^{t+1} = \frac{\ema^{t+1}}{1 - \beta^{t+1}}\, .
\]
The hyperparameter $0 \leq \beta \leq 1$ is the decay factor for the pre-conditioned gradient EMA and $\emat^t$ corrects the initialization bias of $\ema^t$ at zero. Similarly, for the the step-size scale hyperparameter $s$, we have
\[
\nabla_{s} L(\w^{t}|\, \X) = -\nabla_{\w} L(\w^{t}|\, \X) \cdot \bnu^t\, .
\]
Thus, applying the EGU updates results in
\begin{equation}
\label{eq:scale-up}
    s^{t+1} = s^t \exp\big(\gamma_s\,\g^t \cdot \bnu^{t} \big)\, ,
\end{equation}
where $\gamma_s \geq 0$ is the scale learning rate hyperparameter. The pseudo-code for the Funneled Stochastic Gradient Descent with Momentum is shown in Algorithm~\ref{alg:funnel}.

\section{Discussion of the Updates}
In this section, we provide an intuitive explanation of the updates~\eqref{eq:gain-up} and \eqref{eq:scale-up} in terms of the gradient flow. We also discuss normalized updates and their connection to previous methods.

\begin{figure*}[t!]
\vspace{-1.0cm}
\begin{center}
\subfigure[]{\includegraphics[width=0.38\textwidth]{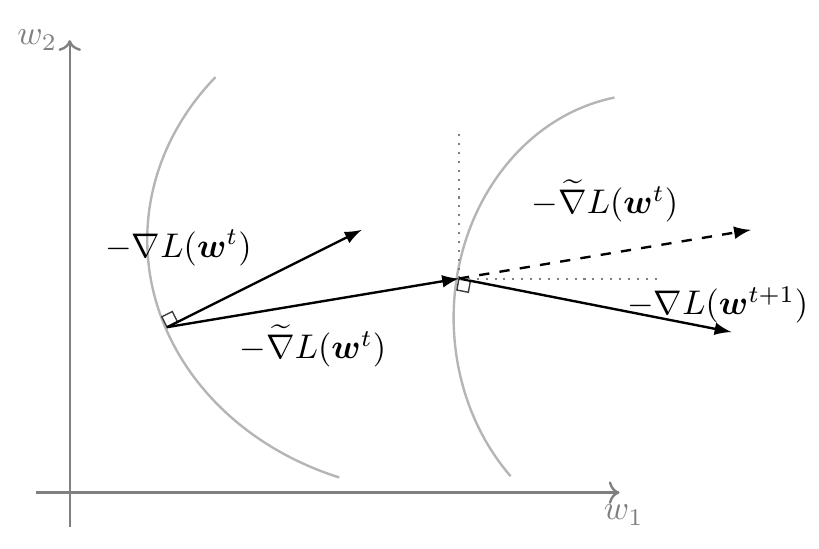}}
    \subfigure[]{\includegraphics[width=0.38\textwidth]{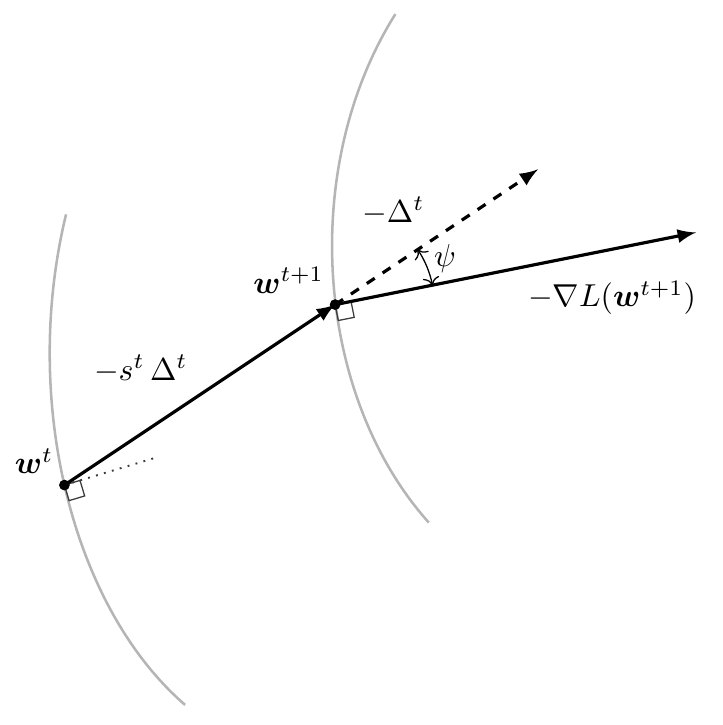}}
    \vspace{-0.25cm}
\end{center}
\vspace{0.8cm}
     \caption{Intuitive explanation of the updates in terms of adjusting the scale and gains based on the gradient flow: (a) adjusting the per-coordinate gain based on the alignments of the gradients in each direction. This requires an independence assumption among different coordinates. (b) Adjusting the step-size scale based on the alignment of the gradient flow. An implicit gradient update corresponds to the case where the previous update step and the new gradient are equal.}\label{fig:flow}
\end{figure*}

\subsection{Coordinate-wise Descent Directions}
Notice that the gain hyperparameters $\p$ in~\eqref{eq:egdd-main} are multiplied element-wise by the pre-conditioned gradient. This is equivalent to introducing a coordinate-wise scale on the pre-conditioned gradients, which implies an independence assumption across the coordinates. In Figure~\ref{fig:flow}(a), we pictorially show the negative gradient $-\g^t = -\nabla_{\w}L(\w^t)$ as well as the negative pre-conditioned gradient $\gt^t = -\partilde_{\w}L(\w^t)$ at $\w^t$. Assuming that we take a step along the negative pre-conditioned gradient, we arrive at the new point $\w^{t+1}$, at which we plot the new negative gradient $-\g^{t+1} = -\nabla_{\w}L(\w^{t+1})$. We also re-draw the negative pre-conditioned gradient $-\gt^t$ at the previous step with a dashed vector. Looking at each coordinate as an independent one-dimensional optimization problem, the descent direction for each coordinate can be obtained by decomposing $-\g^{t+1}$. Combining this with the update~\eqref{eq:gain-pre} implies that those coordinates for which the corresponding component of $-\g^t$ would still have been a \emph{descent direction} at the new point $\w^{t+1}$ are assigned a higher gain value. For the remaining coordinates for which the direction of descent direction switches, the gain values are reduced\footnote{Note that a better approach would be to consider the inter-dependency of all coordinates for updating the gain values. This requires applying a matrix form of gains on the gradients, which in some settings would be drastically more extensive. We consider extensions of our formulation to the gain matrix pre-conditioner and its efficient approximations as a future research direction.}.

\subsection{Adjusting the Discretization Step-size of the Gradient Flow}
When minimizing a loss function $L$ in a full-batch setting, the steepest descent update can be motivated as an iterative procedure that minimizes the loss while remaining close the current value of the parameters in terms of Euclidean distance. This is achieved by minimizing the loss $L$ plus a squared Euclidean divergence as the inertia term,
\begin{equation}
\label{eq:gd-motovation}
\w^{t+1} = \argmin_{\wt \in \RR^d}\, \sfrac{1}{s^t}\,\Vert\!\wt - \w^t\Vert_2^2 + L(\wt)\, .
\end{equation}
Minimizing~\eqref{eq:gd-motovation} directly results in
\begin{align}
\label{eq:gd-implicit}
    \sfrac{1}{s^t}\,\big(\w^{t+1} - \w^t\big) + \nabla_{\w}L(\w^{t+1}) = \zero\, ,\
    \text{\,\,\, i.e.\,\,\, } \w^{t+1} - \w^t = - s\, \nabla_{\w}L(\w^{t+1})\, ,
\end{align}
called the \emph{implicit} gradient descent update ~\cite{pnorm,he2008explicit}. The term implicit implies that the update is motivated using the gradient of the loss function $L$ at a \emph{future} point. In practice, the update is approximated by the \emph{explicit} form in which the gradient at $\w^{t+1}$ is replaced by the gradient at $\w^t$, that is $\nabla_{\w}L(\w^t)$. It has been shown in many cases that the implicit update results in superior convergence than the explicit form~\cite{pnorm,implicit}. The difference between implicit and explicit update stems from the discretization error of the \emph{gradient flow} in continuous-time,
\begin{equation}
\label{eq:gd-flow}
    \dot{\w}(t) = -\nabla_{\w}L(\w(t))\,,
\end{equation}
where $\dot{\w} \coloneqq \frac{\partial \w}{\partial t}$ denotes the time derivative of $\w$. Note that the implicit update~\eqref{eq:gd-implicit} can also be recovered as a \emph{backward Euler} approximation of the gradient flow~\eqref{eq:gd-flow} with \emph{step-size} $s$, while a \emph{forward Euler} approximation results in the explicit update. The difference between the two approximations depends on the \emph{smoothness} of the gradients as well as the step-size $s$. That is, for smooth regions where the change in gradient from $\w^t$ to $\w^{t+1}$ is small, a larger step-size $s$ can be adopted and vice versa. Now consider an update $\Delta^t$ proposed by a given optimizer at step $t$, that is,
\begin{equation}
\label{eq:delta-update}
\w^{t+1} - \w^t = -s^t\, \Delta^t\, .
\end{equation}
This is shown pictorially in Figure~\ref{fig:flow}(b). Note that the update $\Delta^t$ is not necessarily the steepest descent direction and can be generated by any pre-conditioning of the gradient and/or addition of momentum terms internally by the optimizer. However, when $\Delta^t \approx \partial_{\w}L(\w^{t+1})$ implies that the update~\eqref{eq:delta-update} closely approximates the implicit update~\eqref{eq:gd-implicit}. This implies that the gradient of the function $L$ in the neighborhood of $\w^t$ is \emph{smooth} enough such that a larger step-size $\s^{t+1}$ in the next iteration is plausible. This property can be roughly quantified in terms of alignments of the directions by only considering the cosine of the angle $\psi$ between the two vectors, as discussed in the next section.

\begin{figure*}
\vspace{-0.4cm}
{\centering
\begin{minipage}{0.85\linewidth}
\begin{algorithm}[H]
    \centering
    \caption{Funnelled Stochastic Gradient Descent with Momentum}\label{alg:funnel}
    \begin{algorithmic}
        \State \textbf{Input:} Loss function $L$, internal optimizer $\mathcal{D}$, initial parameter $\w^0$, base learning rate $\eta$, pre-conditioned gradient EMA decay factor $\beta$, gain and step-scale learning rate hyperparameters $(\gamma_p, \gamma_s)$\smallskip

        \State $t, \ema^0, \p^0, s^0 \gets 0, \zero, \one, 1$ \Comment{Initialization}
        \While{\,\,$\w^t$ not converged}\,\,
        \State Obtain\, $\g^t = \nabla_{\w} L(\w^t|\, \X^t)$ \Comment{Gradient}
        \State Obtain\, $\gt^t = \partilde_{\w} L(\w^t|\, \X^t)$ from $\mathcal{D}(L, \w^t, \X^t)$ \Comment{Pre-conditioned gradient}
        \State $\p^{t+1} \gets \begin{cases}\p^t\odot\, \exp\big(\gamma_p\, \g^t \odot \emat^{t} \big) & \text{(Unnormalized)}\\[1mm]
        \p^t\odot\, \exp\big(\gamma_p \sign(\g^t) \odot \sign(\ema^t) \big) & \text{(Normalized)}
        \end{cases}$
        \State $s^{t+1} \gets \begin{cases}s^t\exp\Big(\gamma_{s}\,\g^t\cdot \bnu^t\Big)& \text{(Unnormalized)}\\[2mm]
        s^t\exp\Big(\gamma_{s}\,\frac{\g^t}{\Vert\g^t\Vert}\cdot \frac{\bnu^t}{\Vert \bnu^t\Vert}\Big)& \text{(Normalized)}\end{cases}$
        \State $\ema^{t+1} \gets \beta\,\ema^t + (1-\beta)\, \gt^t$
        \State $\bnu^{t+1} \gets \mu\, \bnu^t + \eta\, \big(\p^{t+1}\, \odot\, \gt^t\big)$
        \State $\w^{t+1} \gets \w^t - s^{t+1}\, \bnu^{t+1}$ \Comment{Parameter update}
        \State $t \gets t+1$
        \EndWhile
        \State \textbf{return}\, $\w^t$
    \end{algorithmic}
\end{algorithm}
\end{minipage}
\par
}
\end{figure*}

\subsection{Normalized Updates}

The updates~\eqref{eq:gain-up} and~\eqref{eq:scale-up} are highly dependant on the norm of the gradients at each layer and therefore, may require carefully tuned $(\gamma_p, \gamma_s)$ hyperparameters in each layer. In order to make the updates applicable across different with the same $(\gamma_p, \gamma_s)$, we consider normalized versions of our gain update~\eqref{eq:gain-up} as well as the learning rate scale update~\eqref{eq:scale-up}.

We can approximate the gain update~\eqref{eq:gain-up} by normalizing $\g^t $ and $\ema^{t-1}$ along each coordinate. This corresponds to $\g^t  \oslash \vert \g^t  \vert = \sign(\g^t )$ and $\ema^{t-1} \oslash \vert \ema^{t-1} \vert = \sign(\ema^{t-1})$, i.e., using the signs of the gradient and the EMA term\footnote{Note that the EMA $\ema^{t-1}$ term and the bias corrected version $\emat^{t-1}$ have the same sign.}. This yields the normalized gain update,
\begin{equation}
    \label{eq:eg-gain-norm}
    \g^{t+1} = \g^t\odot\, \exp\big(\gamma_p\, \sign(\g^t) \odot \sign(\emat^t) \big)
\end{equation}

We can also apply a similar normalization to the learning rate scale update~\eqref{eq:scale-up}.  That is, we apply a normalized update by dividing $\g^t $ and $\bnu^{t-1}$ by their $L_2$-norms i.e. only considering the directions and replacing the inner-product by the \emph{cosine similarity},
\begin{equation}
    \label{eq:eg-scale-norm}
    s^{t+1} = s^t\exp\Big(\gamma_{s}\,\frac{\g^t }{\Vert\g^t \Vert}\cdot \frac{\bnu^t}{\Vert \bnu^t\Vert}\Big)\, .
\end{equation}
The normalized update~\eqref{eq:eg-scale-norm} is specially useful for optimizing  multi-layer deep neural networks where the size of the layers vary significantly across the network. The normalization assures that a single hyperparameter $\gamma_s$ can be applied across layers. Note that the learning rate update in~\cite{hypergrad} can be recovered as an approximation of~\eqref{eq:eg-scale-norm}. That is, using the approximation $\exp(x) \approx 1 + x$ yields
\[
s^{t+1} \approx s^t\Big(1 + \gamma_{s}\,\frac{\g^t }{\Vert\g^t \Vert}\cdot \frac{\bnu^t}{\Vert \bnu^t\Vert}\Big)\, .
\]
which is the adaptive update proposed in Hypergradient Descent~\cite{hypergrad}. A similar approximation on the normalized gain updates~\eqref{eq:eg-gain-norm} resembles the multiplicative update form of the DBD method~\cite{dbd}.


\begin{figure}
\vspace{-0.3cm}
     \centering
     \subfigure[]{
         \includegraphics[width=0.22\linewidth]{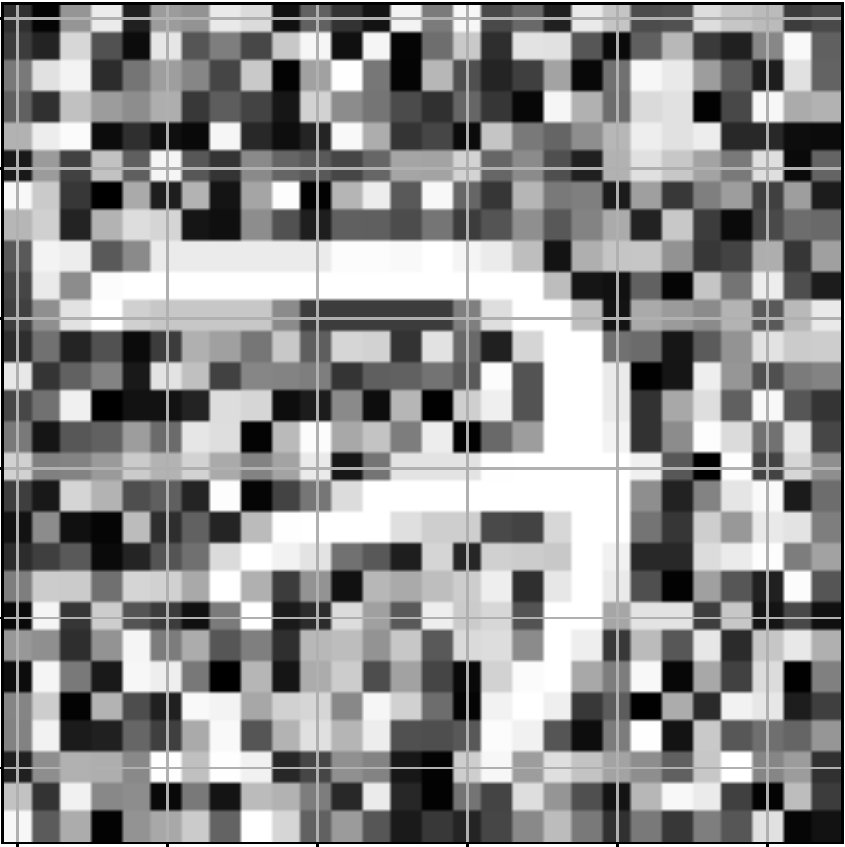}\label{fig:rot0}
         }\,\,
     \subfigure[]{
         \includegraphics[width=0.22\linewidth]{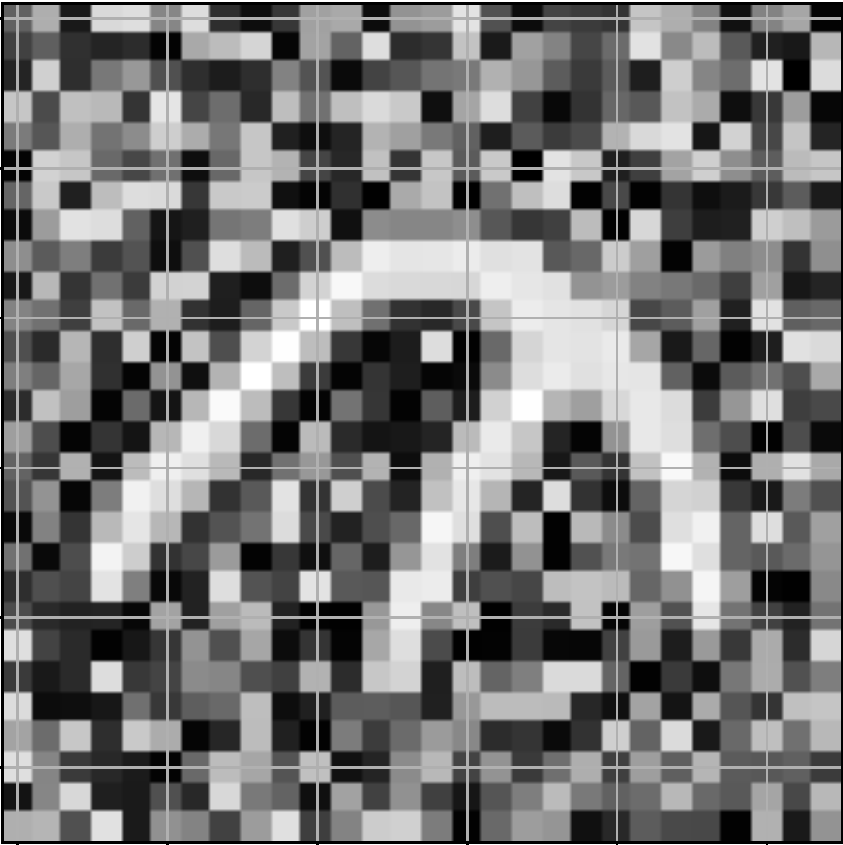}\label{fig:rot45}
     }\,\,
     \subfigure[]{
         \includegraphics[width=0.22\linewidth]{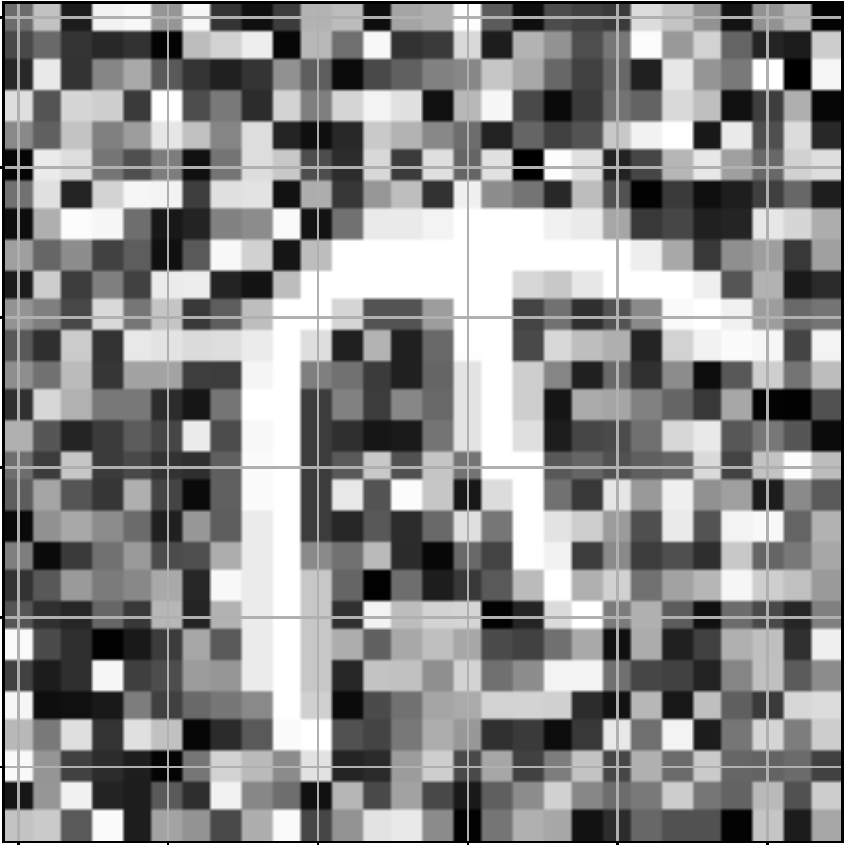}\label{fig:rot90}
     }
        \caption{Samples from the rotated MNIST datasets: random background with (a) 0 degrees, (b) 45 degrees, and (c) 90 degrees rotation.}
        \label{fig:mnist-digits}
\end{figure}

\begin{figure}[t!]
\vspace{-0.4cm}
     \centering
     \subfigure[]{
         \includegraphics[width=0.315\linewidth]{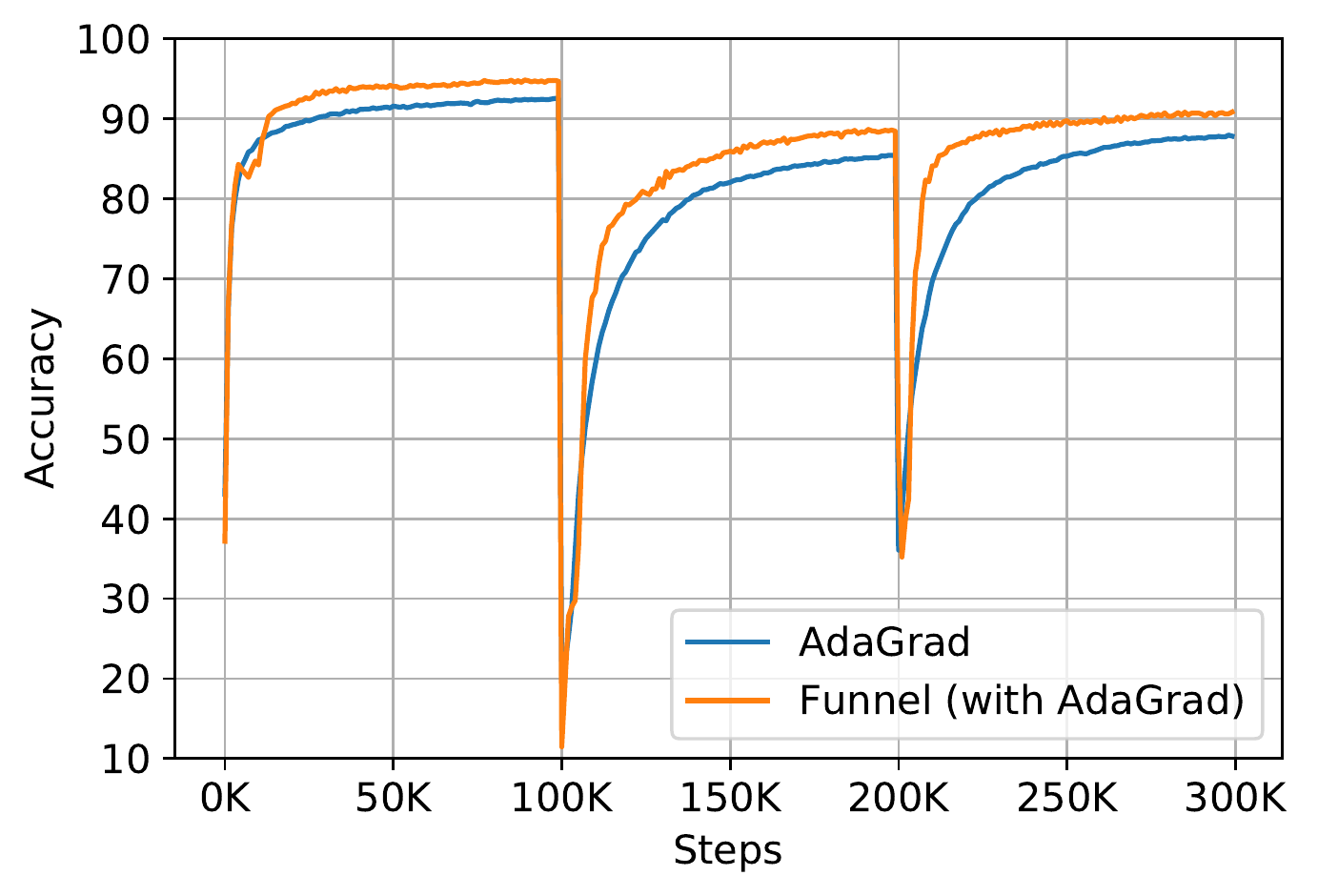}\label{fig:mnistk}
         }
     \subfigure[]{
         \includegraphics[width=0.315\linewidth]{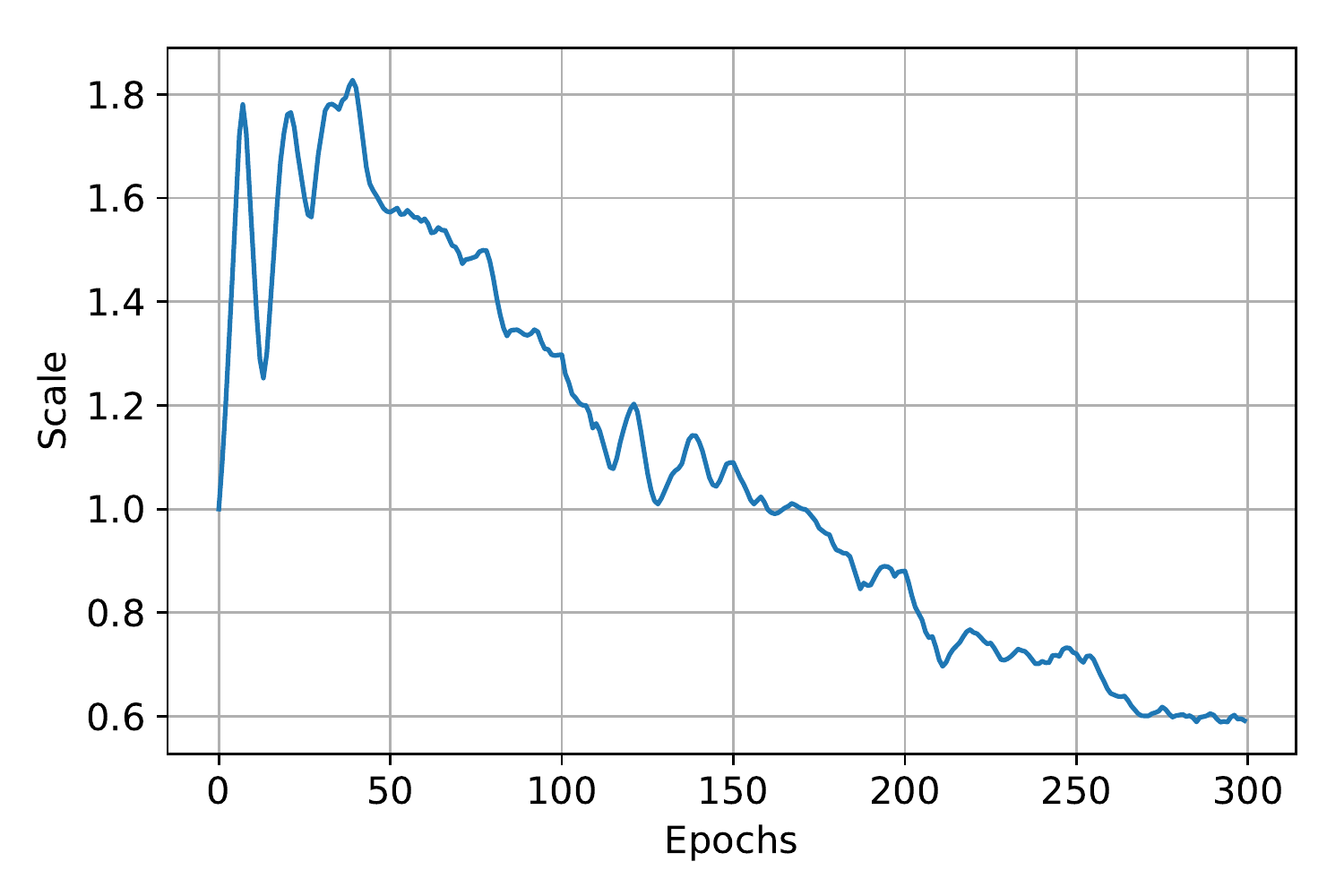}\label{fig:mnist-lr-scales}
     }
     \subfigure[]{
         \includegraphics[width=0.315\linewidth]{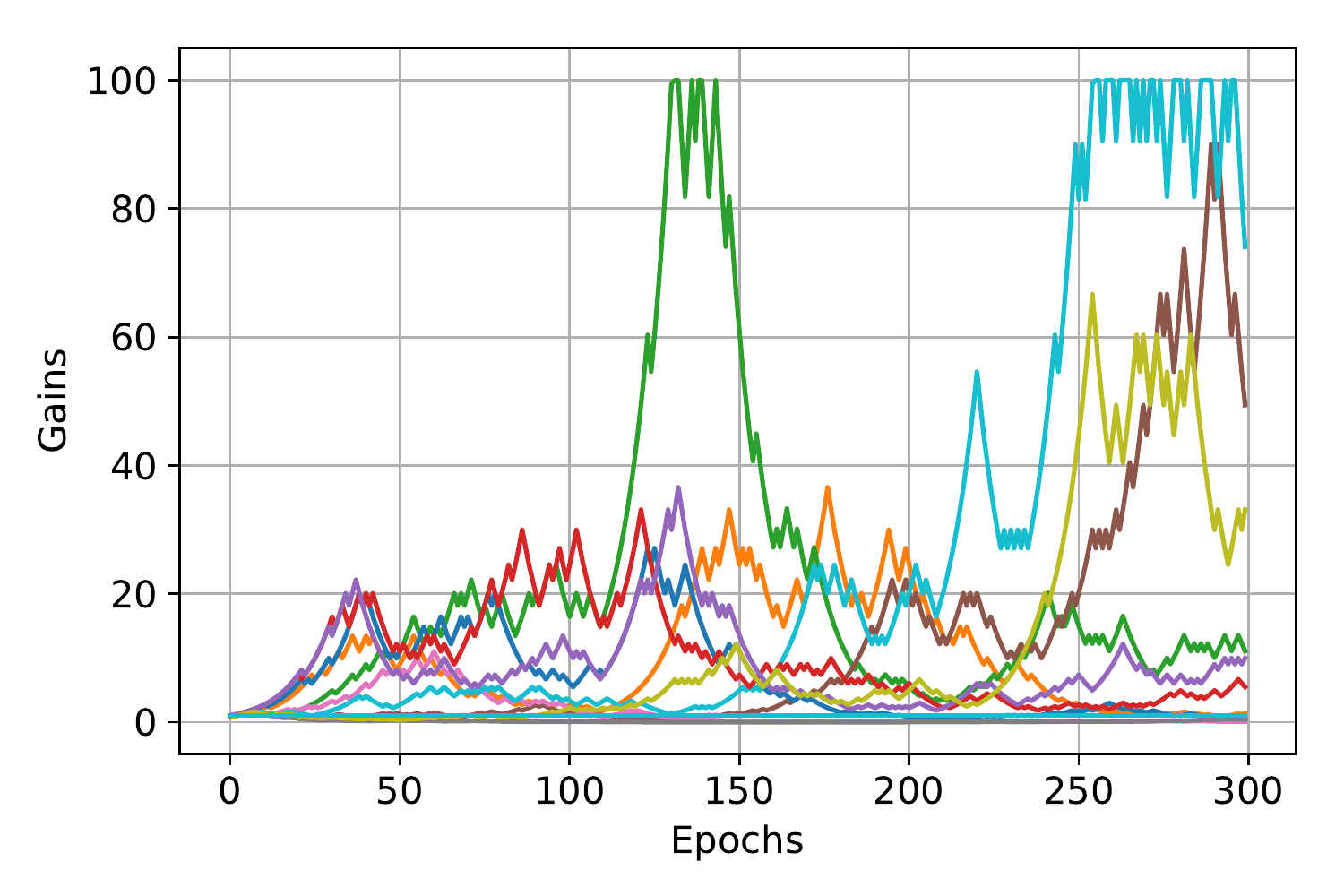}\label{fig:mnist-gains}
     }
        \caption{(a) Top-1 accuracy on shifting MNIST dataset. Every 100k steps (~100 epochs) we shift the datasets (b) Evolution of learning rate scales and (b) Gain values for some of the parameters throughout training.}
        \label{fig:mnist-results}
\end{figure}
\begin{figure}[t!]
\vspace{-0.6cm}
     \centering
     \subfigure[]{
         \includegraphics[width=0.45\linewidth]{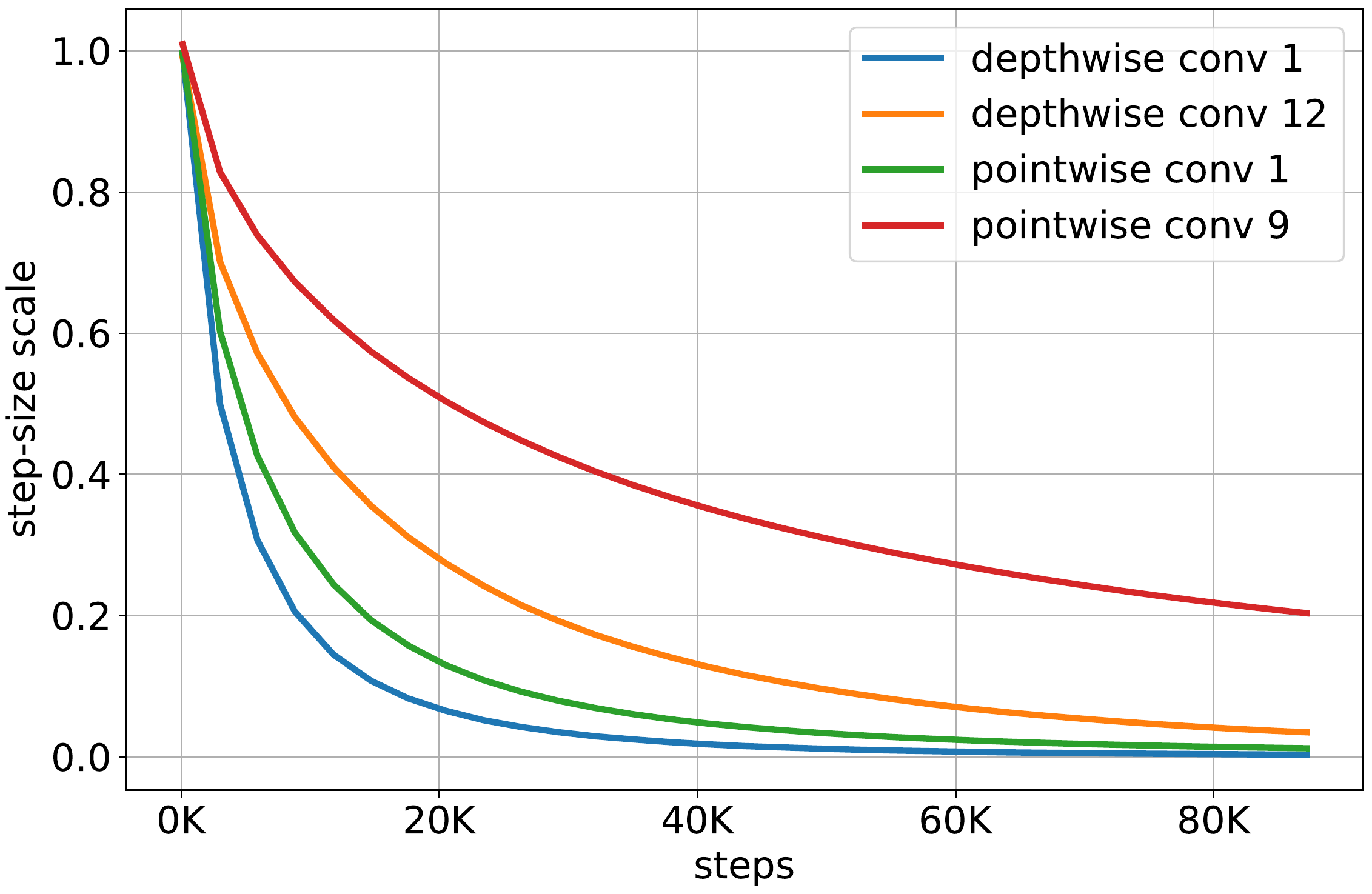}\label{fig:mob_lr_scales}
     }
     \subfigure[]{
         \includegraphics[width=0.45\linewidth]{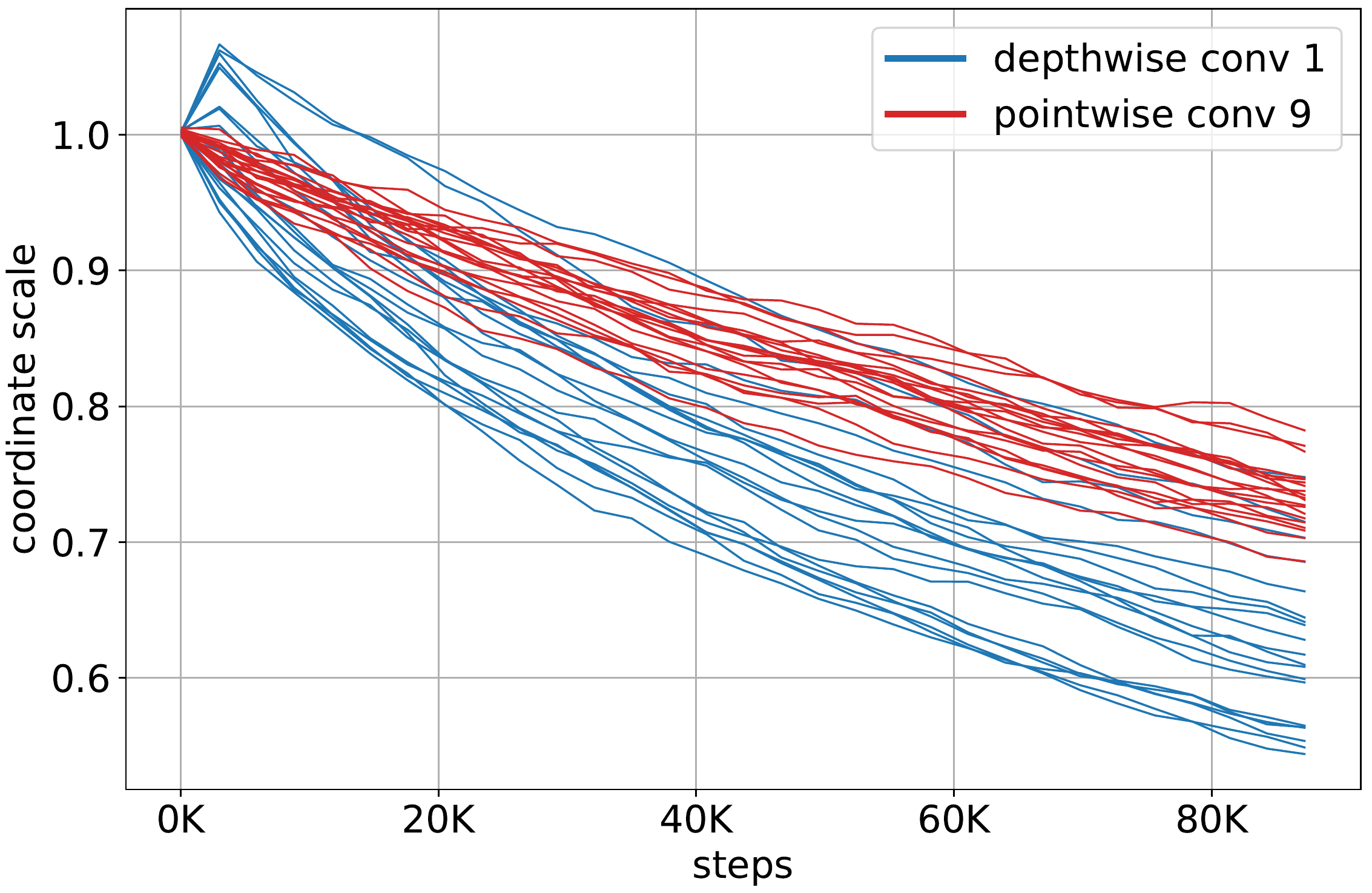}\label{fig:mob_gains}
     }
        \caption{Scale and gain hyperparameters of the MobileNetV1 model: (a) step-size scales for different layers, (b) gains for a subset of coordinates in two different layers. Funnel successfully discovers a decay schedule for all the layers, with a different decay rate. Also, notice that using Funnel the gains vary at a different rate for each layer and each coordinate. Also, some of the gains ramp up initially for the first $\sim$5k steps before starting to decay.}
        \label{fig:mobnet}
        \vspace{-0.4cm}
\end{figure}

\section{Experiments}
\label{sec:experiments}
In this section, we show two cases where our adaptive step-size Funnel method proves to be extremely effective. In the first part of the experiments, we consider the problem of distribution shift in the data during training. For this, we create a synthetic dataset by rotating the MNIST dataset of handwritten digits. We show how Funnel can effectively improve the performance by adapting to the new distribution more rapidly. Next, we consider the setting where we remove the learning rate schedule for training of large-scale models for image classification. We show how the performance of the baseline (as well as the adaptive gradient methods) deteriorates by this change while Funnel can successfully discover a good schedule, adaptively.


\subsection{Distribution Shift}
 Adaptive optimization methods work well on standard datasets where the distribution of the data is fixed. Typical tuning procedure for static datasets involves a decaying learning rate schedule in case of Adam or implicit decay schedule for AdaGrad due to accumulation of gradient statistics. On real world problems where models is training on a stream of freshly arriving data (e.g. click through rate prediction in online advertisement or recommender systems), there is a natural shift in distribution in the dataset over time (e.g. user preferences can change). This requires the optimization method to be adaptive to changing distribution.

To illustrate the advantage of our proposed step-size adaptive mechanism in such cases, we simulate distribution shift on the MNIST dataset of handwritten digits~\cite{mnist} dataset. We split the train, validation, and test into three disjoint sets. For each set, we replace pixel values less than 1e-2 with a random uniform from $[0, 1]$ and rotate the images by 0, 45, and 90 degrees each as shown in Figure~\ref{fig:mnist-digits}. We train a logistic regression model with AdaGrad~\cite{adagrad} as well with Funnel (with AdaGrad as the internal optimizer) at batch size 10k, for 100 epochs for each set sequentially (90 degrees first, followed by 0 degrees, and finally 45 degree). Results are presented in Figure~\ref{fig:mnist-results}. Results indicate that AdaGrad's performance is deteriorated when we switch the training set where as with Funnel higher top-1 accuracy is obtained. We also show the evolution of the learning rate scale, and the per-coordinate gains throughout the training and find the optimization method is able to adjust these hyperparameters in a data dependent way (see Figure~\ref{fig:mnist-lr-scales} and Figure~\ref{fig:mnist-gains}).

\subsection{Adaptive Learning Rate Schedule}
In this section, we conduct experiments on large-scale convolutional neural networks where the baseline model is trained with a highly tuned learning rate schedule. We compare the performance of different optimizer on the same network when the learning rate schedule is removed. We For each experiment, we tune the remaining hyperparameters of the optimizers independently. We also repeat each experiment for the best tuning 5 times and average the results. We consider the following models for the experiments: 1) MobileNetV1 model on CIFAR-10 dataset, and 2) ResNet50 model on the ImageNet dataset.

\subsubsection{MobileNetV1 on CIFAR10 Dataset}
For the MobileNetV1 trained on the CIFAR10 model, we consider the SGD Momentum optimizer as the baseline optimizer. The baseline learning rate schedule consists of two staircase decays without any initial ramp-ups. We train the baseline model for 150k steps with a batch size of 50.

Next, we remove the learning rate schedule and train the model using the SGD Momentum optimizer as well as the Funneled SGD Momentum. For the vanilla Momentum, we use the same learning rate and momentum values as the baseline. We also use the same learning rate and momentum values for Funnel and tune the gain and scale learning rate values in the range $[10^{-5},\, 10^{-3}]$. We set $\beta=0.9$ for the gradient EMA. We also allow the gains and the scale to vary in the range $[0,\, 10^3]$. We repeat each experiment 5 times for 150k iterations and using the same batch size of 50. The best performance for Funnel is achieved with $(\gamma_p, \gamma_s) = (10^{-4}, 10^{-3})$. The results are shown in Table~\ref{tab:mob}. As can be seen from the table, the top-1 accuracy of the baseline Momentum model drops by around 5\% by removing the learning rate schedule. However, our Funnel method can match the baseline performance without using a learning rate schedule. We plot the scale and gain values for a subset of the layers in Figure~\ref{fig:mobnet}. In Figure~\ref{fig:mob_lr_scales}, we plot the scales for 2 depthwise and 2 pointwise convolutional layers. As can be seen from the figure, Funnel is able to recover a decay schedule for all the layers, with a different decay rate. In Figure~\ref{fig:mob_gains}, we show a gains for a subset of coordinates for a depthwise as well as a pointwise convolutional layer. Notice that the gains vary at a different rate for each layer and each coordinate. Also, some of the gains ramp up initially for the first $\sim$5k steps before starting to decay.

\subsubsection{ResNet50 on ImageNet}
We also consider the ResNet50 model trained on the ImageNet dataset. The baseline optimizer corresponds to an SGD Momentum with an initial ramp-up followed by a stair-case decay learning rate schedule. We train the model for 100 epochs using a batch size of 4096. Next, we retrain the model using SGD Momentum, AdaGrad-EMA, and Funneled SGD Momentum optimizers while removing the learning rate schedule. The reason for choosing AdaGrad is the fact the the effective learning rate is naturally decayed for this optimizer, thus mimicking a decaying schedule. Note that AdaGrad-EMA is a slight variant of AdaGrad where we apply an EMA on the pre-conditioned gradients. The original formulation of AdaGrad performs poorly in this setting. We similarly do a hyperparameter search for all the models. The best performing learning rate for AdaGrad-EMA is achieved at $10^{-3}$. For Funnel, we use the original learning rate as the baseline ($\eta = 0.1$) and set $(\gamma_p, \gamma_s) = (10^{-4}, 5\times 10^{-3})$. The results are shown in Table~\ref{tab:resnet}. As can be seen from the table, Funnel achieves the best performance among the other optimizers when the learning rate schedule is removed.

\begin{table}[t!]
\vspace{-0.4cm}
\caption{MobileNetV1 top-1 and top-5 accuracy on the CIFAR-10 dataset: the top baseline result is obtained using a learning rate schedule. We compare the performance of different optimizers when the learning rate schedule is removed.}
\label{tab:mob}
\begin{center}
\resizebox{0.8\textwidth}{!}{
\begin{tabular}{lcc}
\toprule
Method &  Top-1 Test Accuracy & Top-5 Test Accuracy\\
\midrule
Momentum (with lr-schedule) & 
$89.51 \pm 0.18$ & $99.35 \pm 0.05$\\
\midrule
Momentum (without lr-schedule) & 
$84.14 \pm 0.49$ & $98.93 \pm 0.35$\\
Funneled Momentum (without lr-schedule) & 
$\mathbf{89.61 \pm 0.28}$ & $\mathbf{98.94 \pm 0.26}$\\
\bottomrule
\end{tabular}
}
\end{center}
\vspace{-0.4cm}
\end{table}

\begin{table}[t!]

\caption{ResNet50 top-1 and top-5 accuracy on the ImageNet dataset: the top baseline result is obtained using a learning rate schedule. We compare the performance of different optimizers when the learning rate schedule is removed.}
\label{tab:resnet}
\begin{center}
\resizebox{0.8\textwidth}{!}{
\begin{tabular}{lcc}
\toprule
Method &  Top-1 Test Accuracy & Top-5 Test Accuracy\\
\midrule
Momentum (with lr-schedule) & 
$76.57 \pm 0.16$ & $93.21 \pm 0.04$\\
\midrule
Momentum (without lr-schedule) & 
$53.80 \pm 0.86$ & $78.67 \pm 0.74$\\
AdaGrad-EMA (without lr-schedule) & 
$70.70 \pm 0.24$ & $89.74 \pm 0.17$\\
Funneled Momentum (without lr-schedule) & 
$\mathbf{72.39 \pm 0.09}$ & $\mathbf{90.97 \pm 0.06}$\\
\bottomrule
\end{tabular}
}
\end{center}
\vspace{-0.4cm}
\end{table}

\section{Conclusion and Future Work}
We provided an adaptive method for unifying the existing ideas in the domain of learning rate adaptation in a more rigorous manner. This is done by introducing a per-coordinate gain as well as an overall step-size scale and updating these hyperparameter using the well-known unnormalized exponentiated gradient updates. Our meta algorithm can easily adapt to many widely used optimizers, without special modification. We present very promising experimental results for our new adaptive method, e.g. for adapting to distribution shift and finding effective learning rate schedules.


As long-term goal is to also replace the common gradient descent based optimizers by updates from the exponentiated gradient family. For this, a more rigorous study with a focus on large-scale applications is needed. The advantage of using EG updates is that for this family many tools are already available for handling distribution shifts in the data~\cite{herbsterwarmuth,bousquetwarmuth}.

\bibliographystyle{plain}
\bibliography{refs}

\end{document}